\documentclass[letterpaper, 10 pt, conference]{ieeeconf}  

\IEEEoverridecommandlockouts                              

\makeatletter
\let\NAT@parse\undefined
\makeatother
\usepackage[numbers,sort&compress]{natbib}

\usepackage[pdftex]{graphicx}
\usepackage{amsmath}
\usepackage{amssymb}  
\usepackage{subfigure}
\usepackage{subcaption}
\usepackage{multirow}
\usepackage{array,booktabs}
\usepackage{diagbox}
\usepackage{balance}
\usepackage{xspace}
\usepackage{algorithm}
\usepackage{algpseudocode}
\usepackage{adjustbox}
\usepackage{romannum}

\usepackage[final]{hyperref}
\hypersetup{
 colorlinks=true,
 linkcolor=blue,
 filecolor=magenta,
 urlcolor=blue,
 citecolor=black
}
\usepackage{cleveref}
\crefname{figure}{Fig.}{Figs.}
\Crefname{figure}{Fig.}{Figs.}
\usepackage[font=small]{caption}
\usepackage{rpm_SIunits}
\usepackage{rpm_acronyms}
\usepackage{rpm_math}
\usepackage{rpm_misc}
\usepackage{bbm}

\usepackage{textcomp}

\usepackage{soul,color}
\usepackage{lipsum}
\usepackage{dsfont}

\usepackage{xcolor}
\usepackage{colortbl}

\usepackage{amssymb}
\usepackage{pifont}
\newcommand{\cmark}{\ding{51}}%
\newcommand{\xmark}{\ding{55}}%
\usepackage{tikz} 
\usepackage{subcaption}
\usepackage{caption}
\title{\LARGE \bf TreeLoc: 6-DoF LiDAR Global Localization in Forests\\ via Inter-Tree Geometric Matching
}     

\author{
\makebox[\textwidth][c]{%
    Minwoo Jung${}^{1}$, 
    Nived Chebrolu${}^{2}$, 
    Lucas Carvalho de Lima${}^{3}$, 
    Haedam Oh${}^{2}$, 
    Maurice Fallon${}^{2}$ and 
    Ayoung Kim${}^{1*}$%
}%
\thanks{$^\dagger$This work was supported by the National Research Foundation of Korea (NRF) grant funded by the Korea government (MSIT) (No. RS-2024-00461409); the Technology Innovation Program (1415187329, 20024355) funded by the Ministry of Trade, Industry \& Energy (MOTIE, Korea); the Horizon Europe project DigiForest (101070405); and EPSRC Project Mobile Robotic Inspector (EP/Z531212/1). For the purpose of open access, the author has applied a Creative Commons Attribution (CC BY) license to any Author Accepted Manuscript version arising.}%
\thanks{$^{1}$M. Jung and A. Kim are with the Dept. of Mechanical Engineering, SNU, Seoul, S. Korea {\tt\small [moonshot, ayoungk]@snu.ac.kr}}%
\thanks{$^{2}$N. Chebrolu, H. Oh and M. Fallon are with the Oxford Robotics Institute, University of Oxford, UK {\tt\small [nivedc, haedam, mfallon]@robots.ox.ac.uk}}%
\thanks{$^{3}$L. Carvalho de Lima is with the CSIRO Robotics, DATA61, CSIRO and the School of Electrical Engineering and Computer Science, The University of Queensland (UQ), Brisbane, Australia {\tt\small Lucas.Carvalhodelima@data61.csiro.au}}%
}

\begin{document}

\maketitle
\thispagestyle{empty}
\pagestyle{empty}

\begin{abstract}
Reliable localization is crucial for navigation in forests, where GPS is often degraded and LiDAR measurements are repetitive, occluded, and structurally complex.
These conditions weaken the assumptions of traditional urban-centric localization methods, which assume that consistent features arise from unique structural patterns, necessitating forest-centric solutions to achieve robustness in these environments. 
To address these challenges, we propose TreeLoc, a LiDAR-based global localization framework for forests that handles place recognition and 6-DoF pose estimation. 
We represent scenes using tree stems and their \ac{DBH}, which are aligned to a common reference frame via their axes and summarized using the tree distribution histogram (TDH) for coarse matching, followed by fine matching with a 2D triangle descriptor. 
Finally, pose estimation is achieved through a two-step geometric verification. 
On diverse forest benchmarks, TreeLoc outperforms baselines, achieving precise localization. 
Ablation studies validate the contribution of each component. 
We also propose applications for long-term forest management using descriptors from a compact global tree database.
TreeLoc is open-sourced for the robotics community at \textcolor{blue}{https://github.com/minwoo0611/TreeLoc}.

\end{abstract}

\section{Introduction}
\label{sec:intro}
\begin{figure}[!t]
    \centering
    \includegraphics[width=.93\columnwidth]{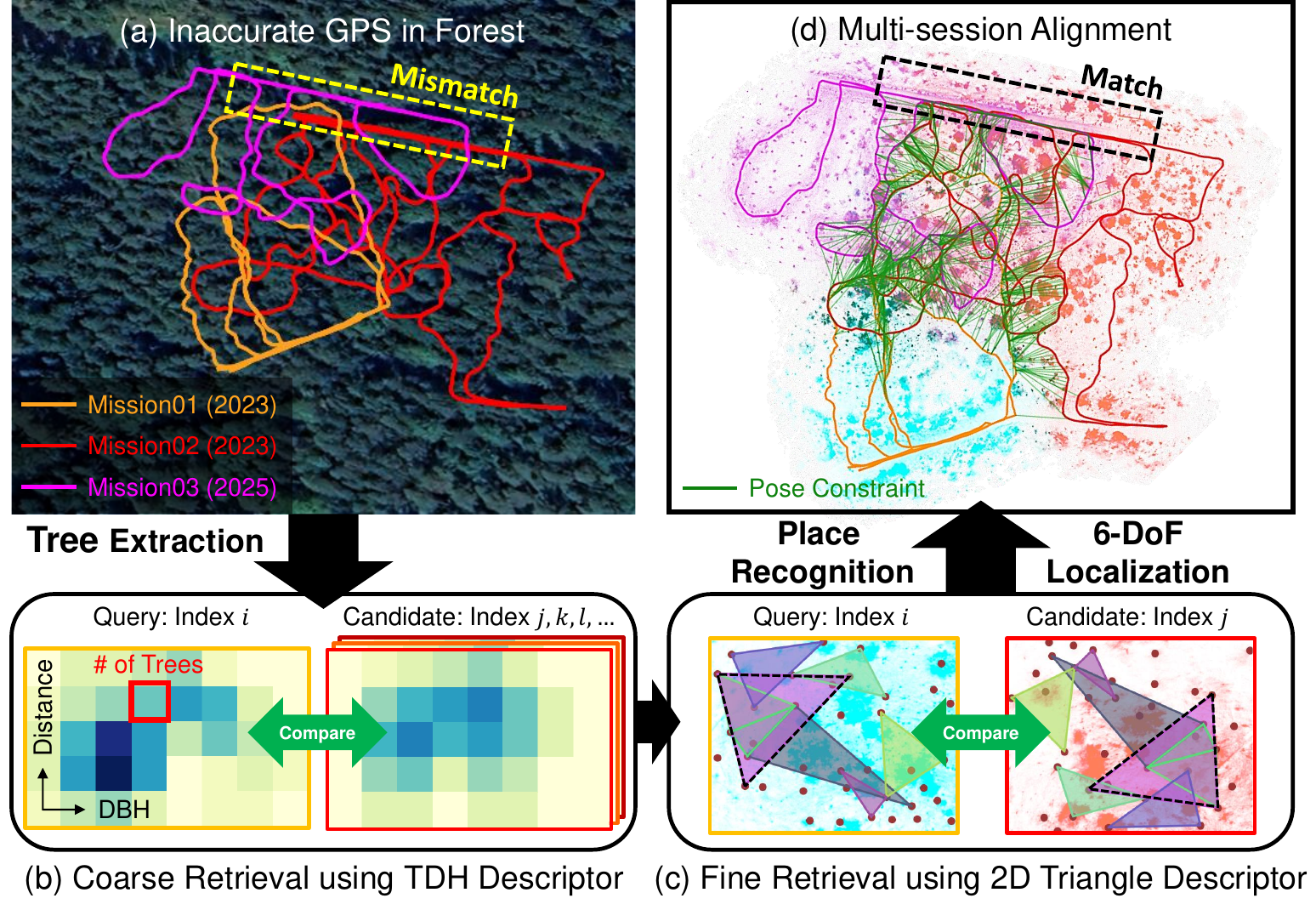}
    \vspace{-1.5mm}
    \caption{(a) GPS bias in forests results in misalignment of different LiDAR SLAM trajectories. (b) TreeLoc performs coarse retrieval using TDH descriptors summarizing tree counts by location and DBH. (c) Fine retrieval with 2D triangle descriptors from tree centers (brown circles); eight matched triangles are highlighted in the same colors. (d) A precise 6-DoF pose enables successful multi-session alignment across datasets captured over different years.}
    \label{fig:Fig1}
    \vspace{-7.3mm}
\end{figure}

Reliable navigation in forests is essential for forestry applications \cite{mattamala2025building} such as digital forest inventory and ecological monitoring.
However, complex scenes and unreliable GPS signals cause substantial challenges, leading to inaccurate localization as shown in \figref{fig:Fig1}(a).
In these situations, LiDAR-based global localization \cite{yin2024survey}, which uses pre-built maps to estimate global poses, is often employed due to its effectiveness in GPS-denied environments.
Nevertheless, these approaches~\cite{knights2023wild} still struggle in forests due to ambiguous structures, seasonal changes, and the storage burden of maintaining dense point cloud maps over long periods.

Despite the need for robust solutions, forest localization is less well studied, with most research~\cite{jung2024helipr} focusing on structured urban environments. 
While existing urban localization methods have been considered for forests, descriptors designed for such environments often suffer from reduced performance in this domain.
For instance, global descriptors \cite{kim2021scan, xu2023ring++} tend to become ambiguous in dense foliage, while local descriptors \cite{gupta2024effectively, yuan2024btc} suffer from keypoint inconsistency as the contours of a trunk are hidden under heavy occlusion.
These limitations highlight the need for a forest-specific approach that uses stable and interpretable object-level features, such as tree stems, which are standard elements in digital forest inventories.
These compact representations are more storage-efficient and easier to maintain than raw point clouds.

Recently, forest localization has achieved more attention with the development of dedicated datasets \cite{ oh2024evaluation, liu2024botanicgarden, malladi2025icra} and applications \cite{freissmuth2024online, wielgosz2024segmentanytree}. 
Unfortunately, many approaches rely on learned features \cite{shen2025forestlpr, de2024online} or segment-level inputs \cite{tinchev2019learning}, requiring intensive computation that limits real-time deployment on resource-constrained hardware. 
Moreover, segment-based methods \cite{tinchev2018seeing, tinchev2019learning} generate one descriptor per point cloud-level cluster, often capturing non-structural elements like twigs and lacking inter-tree geometry. 
These descriptors are also sensitive to appearance changes (i.e., foliage growth over time) and require storing large point clouds, making them unsuitable for long-term management.
To address these issues, we adopt a learning-free, stem-centric representation that models each tree with a small set of parameters: 3D axis, base location, and \acf{DBH}. These parameters are actually the required elements for building digital forest inventories, which foresters leverage to develop harvesting and maintenance plans. By using such inventories, we aim to exploit the spatial arrangement of nearby trees to define geometry signatures suitable for robust localization.

Our proposed global localization method for forests, TreeLoc, is motivated by a perhaps obvious observation: although changes in seasonal foliage growth are significant, the spatial location of the trees themselves does not change, making stem-based features well-suited for persistent mapping.
The pipeline first aligns the scene to a consistent local frame using tree orientations, enabling robust 2D inter-tree geometry despite viewpoint variations.
It then performs a two-stage search: (i) a tree distribution histogram (TDH) encodes tree counts by DBH and distance into a global descriptor for fast retrieval; (ii) 2D triangle descriptors from tree centers refine candidates by capturing inter-tree geometry.
These descriptors and geometric cues, like tree axes and base heights, enable 6-DoF pose estimation without additional feature extraction.
Our main contributions are as follows:

\begin{itemize}
\item A global localization method for forests that leverages robust and repeatable parametric tree features and inter-tree geometry to estimate accurate 6-DoF poses with a compact, learning-free representation.
\item A dual descriptor pipeline: TDH for coarse retrieval and a 2D triangle descriptor using tree centers for fine verification, both stabilized by tree-based scene alignment for rapid candidate retrieval and reliable matching.
\item 6-DoF pose estimation without additional features beyond place recognition, relying solely on 2D triangle and tree geometry to minimize computational overhead.
\item Strong performance over \ac{SOTA} methods, with accuracy comparable to learning-based methods. It enables fast localization within \unit{50}{ms}, and supports multi-session map maintenance with a compact representation, 3-4 orders of magnitude smaller than methods using dense point clouds or descriptor maps.

\end{itemize}

\section{Related Work}
\label{sec:relatedwork}
\subsection{Forestry Applications with LiDAR}
LiDAR has become a key tool in forest inventory \cite{mattamala2025building}, enabling individual tree segmentation and attribute measurement, such as height and DBH, using algorithmic \cite{freissmuth2024online} or deep-learning methods \cite{wielgosz2024segmentanytree}. Public forest datasets \cite{knights2023wild, oh2024evaluation, liu2024botanicgarden, malladi2025icra} have further accelerated progress by providing standardized benchmarks. Recently, research on forest-specific localization has grown, leveraging learned descriptors \cite{shen2025forestlpr, de2024online, tinchev2019learning} and segment-level inputs \cite{tinchev2019learning, tinchev2018seeing}. ForestLPR \cite{shen2025forestlpr} extracts multi-channel \ac{BEV} features from height bins but remains sensitive to clutter in the undergrowth and discretization choices. FGLoc6D \cite{de2024online} employs EgoNN \cite{komorowski2021egonn} with semantic supervision for cross-view retrieval, which depends on segmentation masks. 
Segment-based methods such as ESM \cite{tinchev2019learning} and NSM \cite{tinchev2018seeing} exhibit low temporal robustness to seasonal and viewpoint changes, as they include non-tree elements like foliage alongside trees in the segmented point cloud and overlook inter-tree relationships.
Moreover, these learning-based methods often incur computational overhead that hinders real-time deployment on resource-constrained platforms. In contrast, our approach leverages RealtimeTrees \cite{freissmuth2024online} to directly extract key forest inventory components, such as individual trees and DBH, enabling learning-free, stem-centric features for efficient forest localization.

\subsection{LiDAR-based Global Localization}
Global localization is typically divided into two stages: place recognition and pose estimation, both of which rely on global or local descriptors.
Global descriptors \cite{kim2018scan, kim2021scan, kim2024narrowing} compress point clouds into low-dimensional representations for fast retrieval. Scan Context \cite{kim2018scan, kim2021scan} encodes maximum height on a \ac{BEV} grid, while RING++ \cite{xu2023ring++} enhances rotational invariance using the Radon transform. These methods are usually computationally lightweight but struggle in forests, as repetitive and ambiguous structures can cause over- or under-representation of salient cues, resulting in inadequate information for precise pose estimation.

Local descriptor-based methods aim to overcome these limitations by focusing on sparse yet distinctive geometric cues that facilitate fine-grained pose estimation. MapClosure \cite{gupta2024effectively} extracts ORB features \cite{rublee2011orb} from BEV images, enabling feature-based matching. STD \cite{yuan2023std} and BTC \cite{yuan2024btc} build hash tables from triangle descriptors defined by three keypoints, offering rotation and translation invariance in urban environments. RE-TRIP \cite{park2025re} leverages LiDAR reflectivity to enhance the distinctiveness of these keypoints.
However, in the forest, these local approaches face several challenges: structural noise from undergrowth, heavy occlusion caused by tree trunks, and a lack of high-reflectivity objects. These factors degrade keypoint repeatability and descriptor discriminability, leading to failures in both place recognition and pose estimation. To address this, we generate a dual descriptor using tree stems and their \ac{DBH}, enabling robust performance and fast retrieval. Moreover, projecting the tree layout onto a 2D plane ignores vertical ambiguity. Nevertheless, it still supports accurate 6-DoF pose estimation with tree axes and base heights, without requiring an expensive terrain model or DEM (Digital Elevation Model).

\begin{figure}[!t]
    \centering
    \includegraphics[width=0.85\columnwidth]{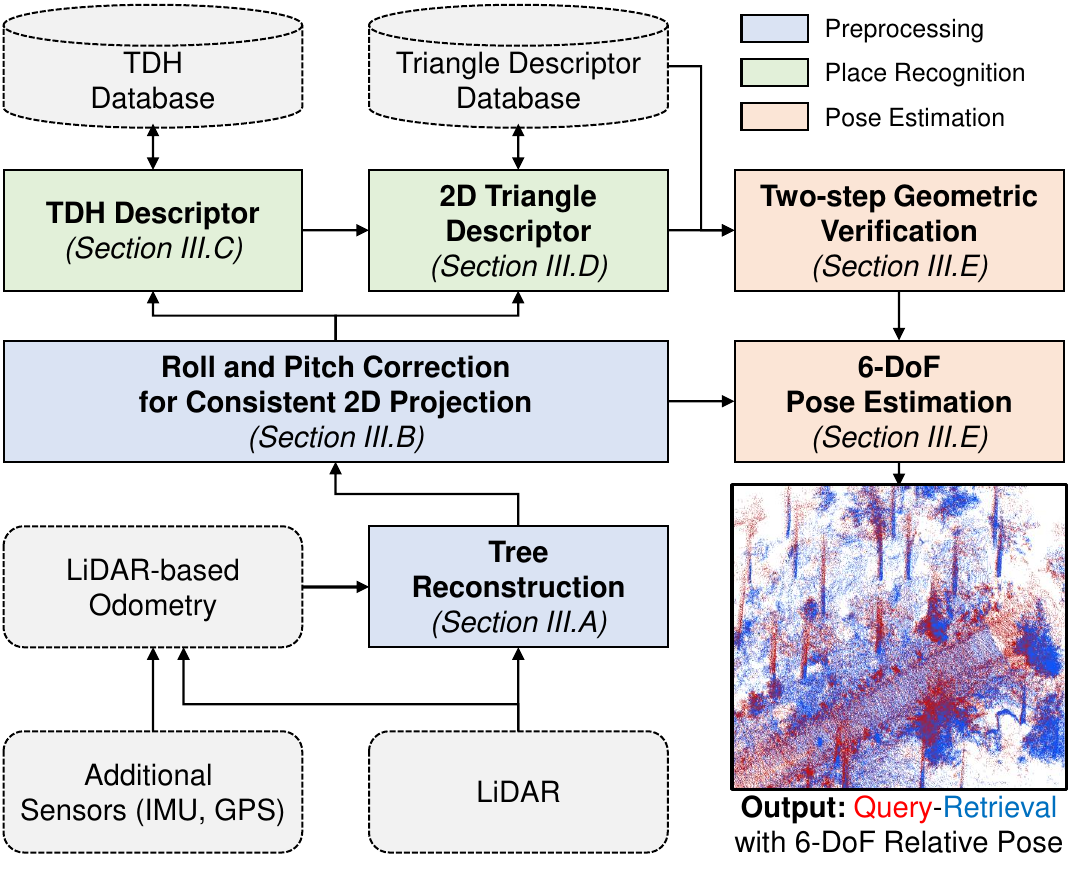}
    \vspace{-2mm}
    \caption{TreeLoc converts LiDAR scans into tree instances. After performing place recognition using two descriptors, it leverages existing data to perform 6-DoF pose estimation. At the end of the pipeline, the point cloud is tightly registered without ICP.}
    \label{fig:Fig2}
    \vspace{-7mm}
\end{figure}

\section{Methodology}
An overview of the TreeLoc pipeline is presented in \figref{fig:Fig2}, comprising three main stages: preprocessing, place recognition via a dual descriptor, and 6-DoF pose estimation.

\subsection{Tree Reconstruction}
\label{sec:reconstruction}
To enable robust tree reconstruction, our pipeline aggregates a number of $k$ LiDAR scans into a single payload to achieve sufficient point density, following recent submap-based LiDAR place recognition \cite{yuan2023std, yuan2024btc}.
Each payload shares a number of $v$ scans with the previous one, ensuring the overlap with adjacent payloads. Payloads are only accumulated when in motion to ensure meaningful aggregation. The poses can be obtained from LiDAR-based SLAM \cite{xu2022fast, wisth2022vilens}, which provides local consistency between scans. 
Let \(\mathcal{P}_u\) denote the payload at index \(u\). We define a window centered at \(t\) as $\mathcal{W}_t = \{ t - \lfloor \frac{s-1}{2} \rfloor, \dots, t + \lfloor \frac{s-1}{2} \rfloor \}$, where a set of $s$ payloads in a sliding window are expressed in the local frame of the index \(t\). The aggregated payloads in the center frame:
\vspace{-0.5mm}
\begin{equation}
\label{eq:agg_submap}
\mathcal{Z}_t \;=\; \bigcup_{u \in \mathcal{W}_t} \mathbf{T}_{t \leftarrow u}\,\mathcal{P}_u ,
\vspace{-0.5mm}
\end{equation}
where \(\mathbf{T}_{t \leftarrow u}\) is the pose from frame \(u\) to the center frame \(t\).

We reconstruct trees from the aggregated payloads \(\mathcal{Z}_t\) using RealtimeTrees \cite{freissmuth2024online}. It first estimates a local terrain surface via a cloth simulation filter and height-normalizes the cloud to suppress slope-induced variance. On the normalized cloud, a Voronoi-inspired clustering method segments vertically coherent stem structures by enforcing spatial compactness in the \(xy\)-plane and continuity along the \(z\)-axis. Stems meeting these criteria are classified as reconstructed trees; others are candidate stems. For each reconstructed tree, the pipeline recovers (i) the 3D tree axis, (ii) the stem center, (iii) \ac{DBH} from a circular fit, and (iv) the base point from tree–terrain intersection. More details are available in RealtimeTrees \cite{freissmuth2024online}. This procedure runs per window \(\mathcal{W}_t\).

We summarize the result at index \(t\) as \(\mathbf{M}_t\), a compact tuple of the pose and a per-tree attribute set, tree database, that serves as the input to TDH and 2D triangle descriptor:
\vspace{-0.7mm}
\begin{equation}
\label{eq:mt_set}
\mathbf{M}_t = (\mathbf{T}_t,\ \mathcal{I}_t),
\qquad
\mathcal{I}_t = \bigl\{\,(\mathbf{A}_j,\ \mathbf{p}'_j,\ d_j)\,\bigr\}_{j=1}^{n_t}.
\vspace{-0.7mm}
\end{equation}
$\mathbf{T}_t$ is the pose at index $t$, $\mathbf{A}_j \in \mathrm{SO}(3)$ is the tree axis orientation, $\mathbf{p}'_j \in \mathbb{R}^3$ is the 3D stem position, combining its center's $x, y$ coordinates ($\mathbf{c}'_j$) with the base height $b'_j$ (the $z$-coordinate of the tree-ground contact), and $d_j$ is the \ac{DBH}. If reconstructed trees are insufficient for generating enough descriptors, we supplement them with candidate stems selected by observation count across payloads.

\subsection{Roll and Pitch Correction for Consistent 2D Projection}
We project the trees onto a 2D plane as a preprocessing step to generate descriptors for place recognition using $\mathbf{M}_t$. Projecting 3D tree structures onto 2D ignores height ambiguity, focusing on stable inter-tree geometry for localization. However, viewpoint-induced roll and pitch variations distort tree center projections on their local $xy$-plane, causing inconsistent center relations.
Basic ground-plane fitting can be used for correction, but it is unreliable on uneven terrain.
To address this, we align scenes to a fixed reference direction using tree axes, estimating a rotation to correct roll and pitch while excluding yaw to ensure consistent 2D projections.
For each tree $j$, we use the unit vector $\mathbf{a}_j$, the third column of $\mathbf{A}_j$. 
The optimal rotation $\mathbf{R}^A_t \in \mathrm{SO}(3)$ is estimated using RANSAC-based robust least-squares optimization, aligning $\{\mathbf{a}_j\}$ to a reference direction.
Any unit vector can be used as direction, but we select $\mathbf{e}_z = (0,0,1)^\top$ for convenience:
\vspace{-1mm}
\begin{equation}
\label{eq:axis_align}
\mathbf{R}^A_t
=\underset{\mathbf{R}\in \mathrm{SO}(3)}{\arg\min}\;
\sum_j \Bigl(1 - \bigl|\mathbf{e}_z^\top \mathbf{R}\mathbf{a}_j\bigr|\Bigr)^2.
\vspace{-1mm}
\end{equation}
$\mathbf{R}^A_t$ aligns each frame to a common reference, ensuring consistent 2D projection across viewpoints. 
Even under an additional rotation $\mathbf{R}_{\text{view}}$, with $\bar{\mathbf{a}}_j=\mathbf{R}_{\text{view}}\mathbf{a}_j$, compensation is achieved with $\mathbf{R}^A_t\mathbf{R}_{\text{view}}^{-1}$ via \eqref{eq:axis_align}.
As $\mathbf{R}^A_t$ encodes roll and pitch relative to the reference frame, it also contributes to 6-DoF pose estimation. Finally, the stem positions are rotated as $\mathbf{p}_j = \mathbf{R}^A_t \mathbf{p}'_j$ and projected onto the plane perpendicular to the reference direction (local $xy$-plane) as
\begin{equation}
\mathbf{c}_j = \Pi_{\perp \mathbf{e}_z}(\mathbf{p}_j) = 
\begin{bmatrix} 1 & 0 & 0 \\ 0 & 1 & 0 \end{bmatrix}
\left( \mathbf{p}_j - (\mathbf{e}_z^\top \mathbf{p}_j)\mathbf{e}_z \right).
\end{equation}

\subsection{Tree Distribution Histogram Descriptor}

We generate a tree distribution histogram (TDH) from the tree database at index \(t\) to encode the spatial layout for fast retrieval. Using 2D centers \(\{\mathbf{c}_j\}\) and their DBH values \(\{d_j\}\), we assign each tree to a radial distance bin (2D distance from the origin) and a DBH bin, as shown in \figref{fig:Fig1}(b).

Radial bins are defined with a fixed resolution \(r_{\text{res}}\) and boundary overlap \(w_{\text{range}}\). DBH spans \([r_{\min}, r_{\max}]\) and is divided into \(n_{\text{sections}}\) bins of width \(w_{\text{dbh}}\) with overlapping intervals. These overlaps mitigate measurement noise and minor pose errors. The resulting histogram \(\mathbf{H}_t \in \mathbb{R}^{n_{\text{spatial}} \times n_{\text{sections}}}\) counts the number of trees in each radial and DBH bin.

To enhance stability, \(\mathbf{H}_t\) is smoothed using a \(2\times2\) uniform filter, leveraging adjacency among neighboring bins. A 40-dimensional descriptor is obtained with \(n_\text{spatial}=5\) and \(n_\text{sections}=8\). For comparison, the chi-square distance between histograms is computed to rank database candidates. Thanks to its compact design, TDH enables fast retrieval, and the top 100 candidates are passed to the fine retrieval stage.

\subsection{2D Triangle Descriptor Using Tree Centers}
To refine TDH matches, we use a 2D triangle descriptor derived from projected centers, which exhibits rotation and translation invariance \cite{yuan2024btc}, as shown in \figref{fig:Fig1}(c). For each scene $\mathcal{I}_s$ with refined centers $\mathcal{C}=\{\mathbf{c}_1,\ldots,\mathbf{c}_{n_t}\}$, each unordered triple $(i,j,k)$ defines
$\mathbf{D}_{ijk}=\{\mathbf{c}_i,\mathbf{c}_j,\mathbf{c}_k,\ell_{ij},\ell_{jk},\ell_{ik},\mathbf{q}_{ijk},s\}$,
where $\ell_{ab}=\|\mathbf{c}_a-\mathbf{c}_b\|_2$ and $\mathbf{q}_{ijk}$ is the triangle centroid.

Permutation invariance is achieved by sorting the side lengths in ascending order and hashing the sorted tuple to form a key:
\(h_{ijk}=\operatorname{hash}(\operatorname{sort}(\ell_{ij},\ell_{jk},\ell_{ik}))\). Each scene stores its keys and descriptors in a hash table \(\mathcal{H}_s\).
The similarity between a query \(Q\) and a candidate \(C\) is defined as:
\(S(Q,C)=\lvert \mathcal{K}_Q \cap \mathcal{K}_C \rvert\),
computed as the number of shared keys, where \(\mathcal{K}_Q\) and \(\mathcal{K}_C\) represent the set of keys extracted from \(\mathcal{H}_Q\) and \(\mathcal{H}_C\). This integer lookup-based comparison is more discriminative than TDH, enabling fine retrieval across hundreds of 2D triangle descriptors. The ten candidates with the highest $S(Q,C)$ are passed to geometric verification.

\begin{figure}[!t]
    \centering
    \includegraphics[width=.95\columnwidth]{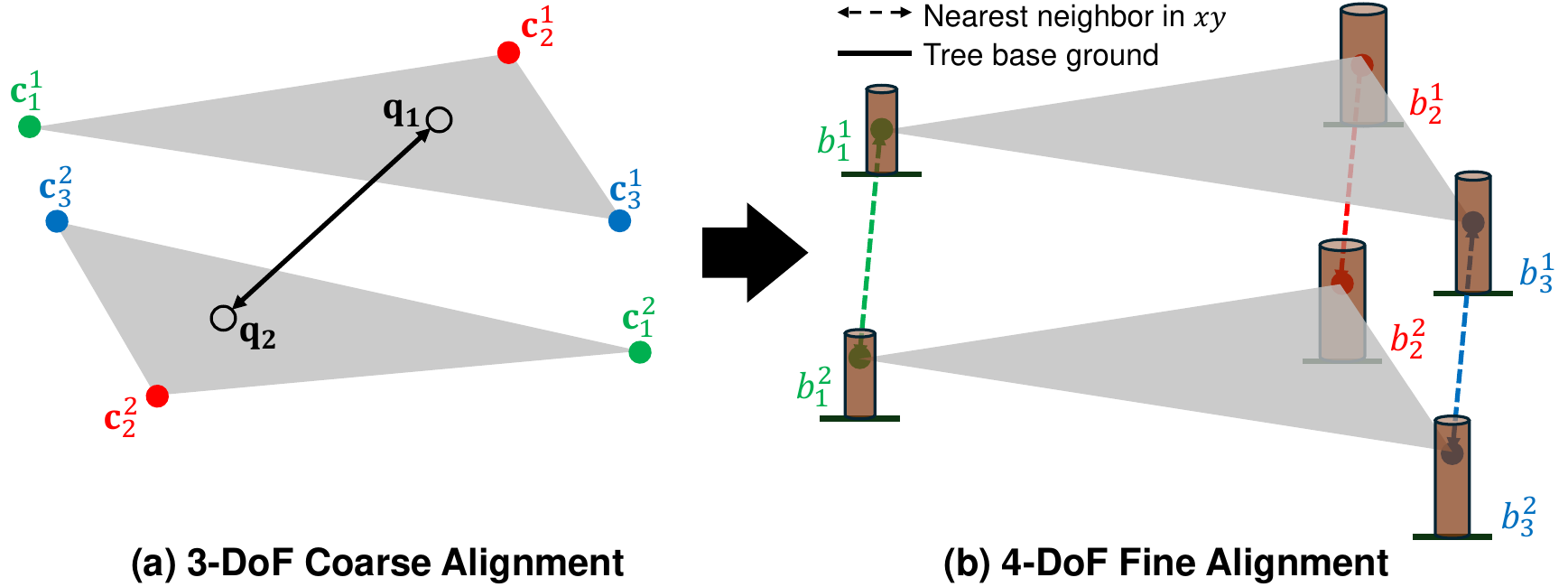}    
    \vspace{-1mm}
    \caption{The coarse stage estimates an initial planar transform from triangle centroids (\(\mathbf{q}\)) matched between the query and candidate scenes. The fine stage refines this planar transform using 2D tree centers (\(\mathbf{c}\)) and base heights (\({b}\)), obtaining the 4-DoF relative pose.}
    \label{fig:align}
    \vspace{-5mm}
\end{figure}

\subsection{Geometric Verification with 6-DoF Localization}
We conduct a two-step geometric verification for each query scene \(\mathcal{I}_Q\) and each of the ten candidate scenes \(\mathcal{I}_C\), calculating a 4-DoF transformation and ranking candidates by geometric consistency, as illustrated in \figref{fig:align}.

\textbf{Initial Planar Alignment:} For each pair of triangles with matching hash keys, we identify a centroid correspondence \((\mathbf{q}^Q_u, \mathbf{q}^C_u)\).
Given \(N\) such pairs, we estimate a planar transform \(\mathbf{T}^c_{C \leftarrow Q} = (\mathbf{R}_c, \mathbf{t}_c) \in \mathrm{SE}(2)\) via SVD-based alignment:
\vspace{-1mm}
\begin{equation}
\label{eq:svd_equation}
\min_{\mathbf{R}_c, \mathbf{t}_c}
\sum_{u=1}^{N}
\bigl\lVert
\mathbf{q}^C_u - (\mathbf{R}_c \mathbf{q}^Q_u + \mathbf{t}_c)
\bigr\rVert^2.
\vspace{-1mm}
\end{equation}

\textbf{Refined Alignment and Vertical Offset:} let \(\mathcal{C}_Q=\{\mathbf{c}_i\}\) and \(\mathcal{C}_C=\{\mathbf{c}_j\}\) be the aligned tree centers, with $\mathcal{B}_Q = \{b_i\}$ and $\mathcal{B}_C = \{b_j\}$ as their base heights.
We transform query centers to \(\widetilde{\mathbf{c}}_i=\mathbf{R}_c\,\mathbf{c}_i+\mathbf{t}_c\). Using a KD-tree built on $\mathcal{C}_C$, we find nearest neighbors for $\widetilde{\mathbf{c}}_i$, retaining pairs with planar distance below $0.4\,\mathrm{m}$ and DBH difference below $0.2\,\mathrm{m}$.
For these pairs, we compute the vertical offset $\Delta z$ by minimizing the base height differences $|b_i - b_j - \Delta{z}|$ via RANSAC.

If at least two pairs remain, we perform a second SVD alignment on the matched centers $\widetilde{\mathbf{c}}_i$ and $\mathbf{c}_j$, yielding a refined transform $\mathbf{T}^f_{C \leftarrow Q} = (\mathbf{R}_f, \mathbf{t}_f) \in \mathrm{SE}(2)$, where $\mathbf{R}_f \in \mathrm{SO}(2)$ and $\mathbf{t}_f \in \mathbb{R}^2$, using \eqref{eq:svd_equation} with $\widetilde{\mathbf{c}}_i$ and $\mathbf{c}_j$ as inputs instead of $\mathbf{q}^Q_u$ and $\mathbf{q}^C_u$.
The 4-DoF transformation in \(\mathrm{SE}(3)\) is composed as \(\mathbf{T}^{4D}_{C \leftarrow Q}=(\mathbf{R}_{4D},\mathbf{t}_{4D})\), where \(\mathbf{R}_{4D}\in \mathrm{SO}(3)\) extends the in-plane rotation \(\mathbf{R}_f\mathbf{R}_c\) with roll and pitch set to zero, and \(\mathbf{t}_{4D}=\bigl[(\mathbf{R}_f \mathbf{t}_c+\mathbf{t}_f)^\top\ \ \Delta z\bigr]^\top\).

\textbf{Geometric Consistency and Ranking:} Let $\mathcal{M}_{Q,C}$ denote the set of matching tree pairs supporting the refined transform, with $|\mathcal{T}_Q|$ and $|\mathcal{T}_C|$ as the number of trees in each scene. The overlap ratio quantifies geometric consistency:
\vspace{-1mm}
\begin{equation}
\mathcal{O}(Q,C) =
\frac{|\mathcal{M}_{Q,C}|}
{|\mathcal{T}_Q| + |\mathcal{T}_C| - |\mathcal{M}_{Q,C}|}.
\vspace{-1mm}
\end{equation}

The ten candidates are ranked by \(\mathcal{O}(Q,C)\).
For the highest scoring candidate, the alignment rotations \(\mathbf{R}^A_Q\) and \(\mathbf{R}^A_C\) are applied to compute a 6-DoF transformation. This yields \(\mathbf{T}^{6D}_{C \leftarrow Q}=(\mathbf{T}^A_C)^{-1}\,\mathbf{T}^{4D}_{C \leftarrow Q}\,\mathbf{T}^A_Q \in \mathrm{SE}(3)\), where \(\mathbf{T}^A_Q\) and \(\mathbf{T}^A_C\) are the pure rotational transformations induced by \(\mathbf{R}^A_Q\) and \(\mathbf{R}^A_C\). Although place recognition is constrained to a 2D plane, the 6-DoF transformation $\mathbf{T}^{6D}_{C \leftarrow Q}$ can still be estimated using only tree information, enabling effective loop closure.

\begin{figure}[!t]
    \centering
    \includegraphics[width=.9\columnwidth]{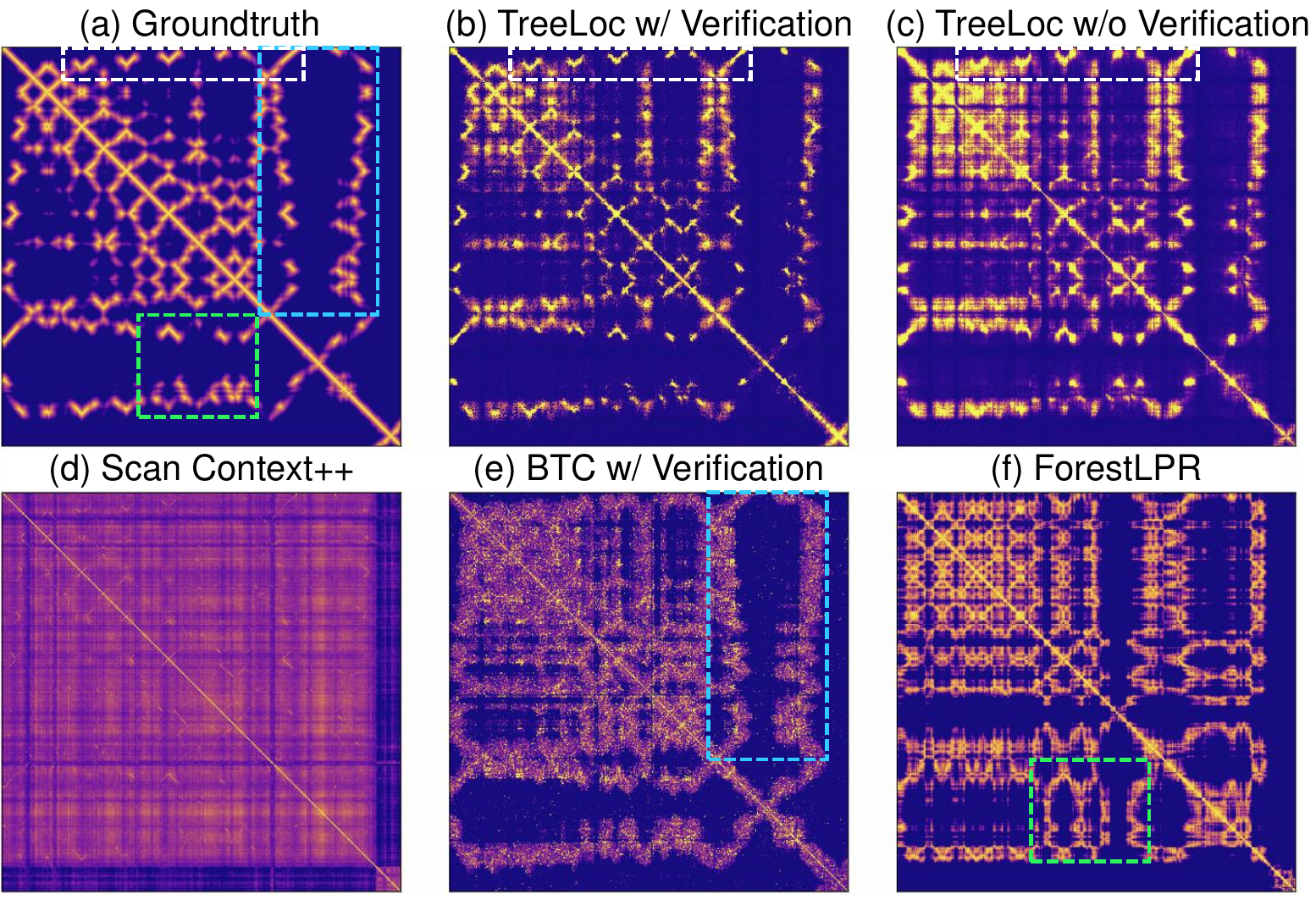}    
    \vspace{-2mm}
    \caption{Feature similarity maps in \texttt{Evo} (warmer colors for higher similarity). TreeLoc focuses similarity in a compact region aligned with the ground truth, enabling clear thresholding, while baselines yield diffuse areas (blue) mixed with false positives (green).}

    \label{fig:feature}
    \vspace{-3mm}
\end{figure}

\begin{figure}[!t]
    \centering
    \includegraphics[width=\columnwidth]{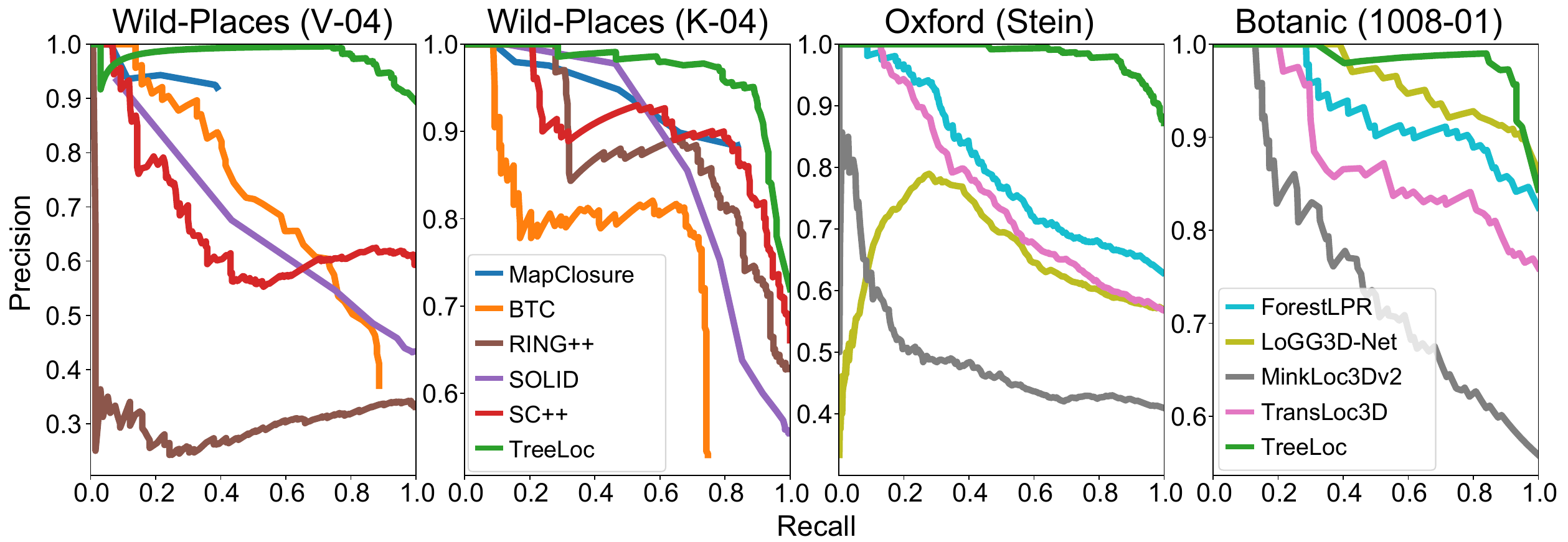}    
    \vspace{-6mm}
    \caption{Precision–Recall curves comparing with algorithmic methods (left) and learning-based methods (right). TreeLoc achieves a higher AUC and maintains high precision at high recall, thereby avoiding the sharp drop in precision observed in baselines.}
    \label{fig:pr_curve}
    \vspace{-7mm}
\end{figure}

\section{experiment}
\label{sec:experiment}


\begin{table*}[!t]
\centering
\caption{Intra-session place recognition with Multiple Forest Dataset (\colorbox[HTML]{def3e6}{\textbf{Bold}}: Best, \colorbox[HTML]{ecf8f1}{Normal}: Second-Best, V: Venman, K: Karawatha)}
\vspace{-1.5mm}
\label{tab:intra}
\resizebox{\textwidth}{!}{%
\begin{tabular}{c|ccc|ccc|ccc|ccc|ccc|ccc}
\toprule
\multicolumn{1}{c|}{} &
  \multicolumn{3}{c|}{Oxford (\texttt{Evo})} &
  \multicolumn{3}{c|}{Oxford (\texttt{Stein})} &
  \multicolumn{3}{c|}{Wild-Places (\texttt{V-03})} &
  \multicolumn{3}{c|}{Wild-Places (\texttt{V-04})} &
  \multicolumn{3}{c|}{Wild-Places (\texttt{K-03})} &
  \multicolumn{3}{c}{Wild-Places (\texttt{K-04})} \\
Method &
  R@1 &
  F1 &
  AUC &
  R@1 &
  F1 &
  AUC &
  R@1 &
  F1 &
  AUC &
  R@1 &
  F1 &
  AUC &
  R@1 &
  F1 &
  AUC &
  R@1 &
  F1 &
  AUC \\ \midrule
SC++       & 0.476 & 0.784 & \cellcolor[HTML]{ecf8f1}0.883 & 0.579 & \cellcolor[HTML]{ecf8f1}0.797 & 0.854 & 0.087 & 0.361 & 0.321 & \cellcolor[HTML]{ecf8f1}0.594 & \cellcolor[HTML]{ecf8f1}0.757 & 0.670  & 0.517  & 0.708  & \cellcolor[HTML]{ecf8f1}0.734 & 0.659 & \cellcolor[HTML]{ecf8f1}0.867 & \cellcolor[HTML]{ecf8f1}0.906 \\
SOLID      & 0.428 & 0.675 & 0.772 & 0.527 & 0.751 & 0.832  & 0.081 & 0.151 & 0.063 & 0.434 & 0.633 & 0.636 & 0.324 & 0.491 & 0.476 & 0.554 & 0.768 & 0.815 \\
RING++     & 0.549 & 0.719 & 0.847 & 0.576 & 0.785 & \cellcolor[HTML]{ecf8f1}0.878 & 0.188 & 0.409 & 0.303 & 0.331 & 0.508 & 0.305 & 0.455 & 0.625 & 0.510 & 0.627 & 0.822 & 0.875 \\ 
BTC        & \cellcolor[HTML]{ecf8f1}0.626 & \cellcolor[HTML]{ecf8f1}0.804 & 0.868 & \cellcolor[HTML]{ecf8f1}0.611 & 0.774 & 0.803 & \cellcolor[HTML]{ecf8f1}0.387 & \cellcolor[HTML]{ecf8f1}0.660 & \cellcolor[HTML]{ecf8f1}0.700 & 0.353 & 0.661 & \cellcolor[HTML]{ecf8f1}0.673 & 0.362 & 0.596 & 0.521 & 0.449 & 0.747 & 0.617  \\
MapClosure & 0.317 & 0.481 & 0.311 & 0.060 & 0.114 & 0.060 & 0.170 & 0.291 & 0.163 & 0.376 & 0.546 & 0.368 & \cellcolor[HTML]{ecf8f1}0.572  & \cellcolor[HTML]{ecf8f1}0.728 & 0.581 & \cellcolor[HTML]{def3e6}\textbf{0.756} & 0.860 & 0.792 \\ 
\cellcolor[HTML]{f3f7fc}\textbf{TreeLoc}       &  \cellcolor[HTML]{def3e6}\textbf{0.907} & \cellcolor[HTML]{def3e6}\textbf{0.966}& \cellcolor[HTML]{def3e6}\textbf{0.992} & \cellcolor[HTML]{def3e6}\textbf{0.871} & \cellcolor[HTML]{def3e6}\textbf{0.945} & 
\cellcolor[HTML]{def3e6}\textbf{0.988} & \cellcolor[HTML]{def3e6}\textbf{0.865} & \cellcolor[HTML]{def3e6}\textbf{0.933} & \cellcolor[HTML]{def3e6}\textbf{0.980} & \cellcolor[HTML]{def3e6}\textbf{0.890} & \cellcolor[HTML]{def3e6}\textbf{0.942} & \cellcolor[HTML]{def3e6}\textbf{0.974} & \cellcolor[HTML]{def3e6}\textbf{0.798} & \cellcolor[HTML]{def3e6}\textbf{0.897} & \cellcolor[HTML]{def3e6}\textbf{0.952} & \cellcolor[HTML]{ecf8f1}0.694 & \cellcolor[HTML]{def3e6}\textbf{0.928} & \cellcolor[HTML]{def3e6}\textbf{0.965} \\ \bottomrule
\end{tabular}%
}
\vspace{-6mm}
\end{table*}

\begin{figure*}[!t]
    \centering
    \begin{minipage}{0.72\textwidth}  
        \centering
        \begin{table}[H]
        \centering
        \caption{Inter-session place recognition performance. Arrows indicate higher/lower is better.}
        \vspace{-1.5mm}
        \label{tab:inter-session-pr}
        \renewcommand{\arraystretch}{1.15}
        \resizebox{\textwidth}{!}{%
        \begin{tabular}{@{}c|cccccc|cccccc@{}}
        \toprule
         & \multicolumn{6}{c|}{{Wild-Places }(\texttt{Venman01-04}, 12 query-database pairs)} & \multicolumn{6}{c}{{Wild-Places} (\texttt{Karawatha01-04}, 12 query-database pairs)} \\ \cline{2-13}
        Method &
        \multicolumn{2}{c}{R@1} &
        \multicolumn{2}{c}{F1} &
        \multicolumn{2}{c|}{AUC} &
        \multicolumn{2}{c}{R@1} &
        \multicolumn{2}{c}{F1} &
        \multicolumn{2}{c}{AUC} \\
         & Mean [$\uparrow$] & $\sigma$ $[\downarrow]$ &
           Mean [$\uparrow$] & $\sigma$ $[\downarrow]$ &
           Mean [$\uparrow$] & $\sigma$ $[\downarrow]$ &
           Mean [$\uparrow$] & $\sigma$ $[\downarrow]$ &
           Mean [$\uparrow$] & $\sigma$ $[\downarrow]$ &
           Mean [$\uparrow$] & $\sigma$ $[\downarrow]$ \\ \midrule
        SC++ &
        \cellcolor[HTML]{ecf8f1}0.594 & 0.152 &
        \cellcolor[HTML]{ecf8f1}0.746 & 0.118 &
        \cellcolor[HTML]{ecf8f1}0.812 & 0.134 &
        \cellcolor[HTML]{ecf8f1}0.579 & 0.138 &
        \cellcolor[HTML]{ecf8f1}0.730 & 0.110 &
        \cellcolor[HTML]{ecf8f1}0.675 & 0.170 \\
        SOLID & 0.358 & 0.177 & 0.511 & 0.177 & 0.514 & 0.156 & 0.373 & 0.131 & 0.557 & 0.134 & 0.542 & 0.128 \\
        RING++ & 0.368 & 0.115 & 0.544 & 0.112 & 0.546 & 0.156 & 0.414 & 0.073 & 0.584 & 0.071 & 0.458 & 0.140 \\
        BTC &
        0.302 & \cellcolor[HTML]{ecf8f1}{0.038} &
        0.500 & \cellcolor[HTML]{ecf8f1}0.039 &
        0.337 & \cellcolor[HTML]{ecf8f1}0.077 &
        0.086 & 0.045 &
        0.243 & 0.125 &
        0.131 & 0.098 \\
        MapClosure &
        0.172 & 0.092 &
        0.286 & 0.125 &
        0.163 & 0.092 &
        0.136 & \cellcolor[HTML]{ecf8f1}0.044 &
        0.237 & \cellcolor[HTML]{ecf8f1}0.068 &
        0.127 & \cellcolor[HTML]{ecf8f1}0.045 \\
        \cellcolor[HTML]{f3f7fc}\textbf{TreeLoc} &
        \cellcolor[HTML]{def3e6}\textbf{0.920} & \cellcolor[HTML]{def3e6}\textbf{0.036} &
        \cellcolor[HTML]{def3e6}\textbf{0.984} & \cellcolor[HTML]{def3e6}\textbf{0.006} &
        \cellcolor[HTML]{def3e6}\textbf{0.993} & \cellcolor[HTML]{def3e6}\textbf{0.005} &
        \cellcolor[HTML]{def3e6}\textbf{0.771} & \cellcolor[HTML]{def3e6}\textbf{0.034} &
        \cellcolor[HTML]{def3e6}\textbf{0.978} & \cellcolor[HTML]{def3e6}\textbf{0.002} &
        \cellcolor[HTML]{def3e6}\textbf{0.989} & \cellcolor[HTML]{def3e6}\textbf{0.001} \\
        \bottomrule
        \end{tabular}%
        }
        \end{table}
        
    \end{minipage}
    \hfill
    \begin{minipage}{0.25\textwidth}  
        \centering
        \includegraphics[trim = 0cm 0cm 0cm 0cm, width=\linewidth]{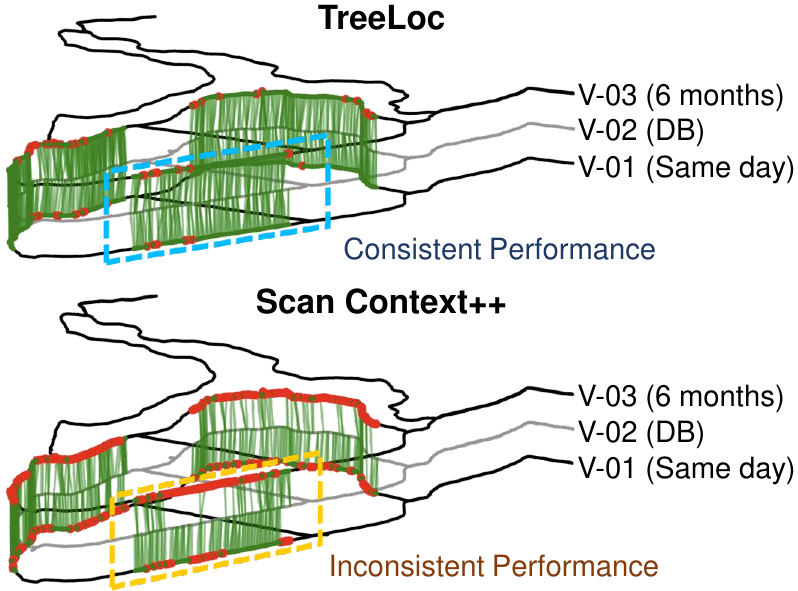}
        \vspace{-8mm}
    \end{minipage}%
    \caption{(Left) Mean and standard deviation ($\sigma$) of metrics across 12 pairs. TreeLoc achieves the highest means with the lowest $\sigma$. (Right) While SC++ succeeds for same-day pairs (\texttt{V-01} and \texttt{V-02}; orange) but fails with temporal discrepancies (\texttt{V-02} and \texttt{V-03}), TreeLoc reliably detects true positives across sessions (\texttt{V-02} and \texttt{V-03}; blue), highlighting its robustness to environmental and temporal variation.}

    \vspace{-6mm}
    \label{fig:consistency}
\end{figure*}

\subsection{Datasets and Evaluation Metrics}
We evaluated TreeLoc on the Oxford Forest Place Recognition dataset \cite{oh2024evaluation} (\texttt{Evo} and \texttt{Stein am Rhein}, where the backpack recording device has strong roll and pitch motion), Wild-Places \cite{knights2023wild} (\texttt{Venman01-04} and \texttt{Karawatha01-04}, natural forests with long trajectories and seasonal changes), and BotanicGarden \cite{liu2024botanicgarden} (\texttt{1008-01}, an urban park).
Payloads were built from trajectories using FAST-LIO2 \cite{xu2022fast} for Oxford and from ground truth for the others.
We reported Recall@1 (R@1), maximum F1 score, and AUC under a \unit{10}{m} threshold for place recognition. For localization, we used Recall@\unit{50}{cm} (R@50), defined as pose estimation with translation error (TE) $\le$ \unit{0.5}{m} and rotation error (RE) $\le 5^\circ$, along with the success rate (SR), the ratio of true positives meeting the R@50 criteria. Errors were computed only for successful cases, following the approach used in BTC \cite{yuan2024btc}.

Since TreeLoc exploits multiple payloads to form a tree-based descriptor, we also provided all baselines with accumulated LiDAR scans within a \unit{60}{m} $\times$ \unit{60}{m} for both query and database inputs to ensure a fair comparison. We compared against three global descriptor-based methods such as Scan Context++ (SC++) \cite{kim2021scan}, SOLID \cite{kim2024narrowing}, and RING++ \cite{xu2023ring++}, two local descriptor-based methods, BTC \cite{yuan2024btc} and MapClosure \cite{gupta2024effectively}, and four learning-based methods, TransLoc3D \cite{xu2021transloc3d}, LoGG3D-Net \cite{vidanapathirana2022logg3d}, MinkLoc3Dv2 \cite{komorowski2022improving}, and ForestLPR \cite{shen2025forestlpr}.

\subsection{Intra-session Place Recognition}
We first evaluated the intra-session place recognition performance of TreeLoc across six sequences in \tabref{tab:intra}. The most recent 50 queries were excluded from the database to avoid trivial matches arising from temporal proximity. Except for \texttt{K-04}, TreeLoc outperformed all \ac{SOTA} baselines in R@1, F1, and AUC, with substantial improvements in F1 and AUC. Global descriptor-based methods (SC++, SOLID, RING++) struggled due to weak distinctiveness in forests, often producing similar descriptors for negative scenes. As shown in \figref{fig:feature}(d), SC++ is unable to clearly distinguish each place, making threshold selection difficult. Low thresholds allowed many false positives, while high thresholds missed true positives. Local descriptor-based methods (BTC, MapClosure) also underperformed due to the repetitive and irregular forest structure, which degrades feature distinctiveness. Although BTC’s similarity map roughly followed the ground truth contour, it produced scattered high-similarity regions near true positives (\figref{fig:feature}(e)), reducing localization accuracy.

In contrast, TreeLoc produces place recognition matches which closely resemble the ground truth, as shown in \figref{fig:feature}(a)-(c). Even without a geometric verification, TreeLoc's descriptor similarity was higher than that of other baselines. While the matched region appeared wider, it was clearly separated from false positives, which simplified the threshold selection process. As shown in \figref{fig:pr_curve}, TreeLoc maintains high precision at high recall, while baselines exhibit a sharp decline in precision. This highlights TreeLoc's robustness in real-world deployment. TreeLoc's performance slightly degrades in \texttt{K-04} due to sparse trees in open spaces, reducing the number of distinct stem-based features for matching. Nevertheless, it outperforms all baselines, confirming the reliability of its tree-based representations.

\subsection{Inter-session Place Recognition}

We evaluated inter-session place recognition on \texttt{Venman} and \texttt{Karawatha} sequences \texttt{01}–\texttt{04} from Wild-Places, using 24 query-database pairs (12 from each region) and following its protocol for comparing repeated traversals in specific test regions. Sequences were collected on the same day, and 6 and 14 months apart, introducing temporal discrepancies and requiring descriptors that preserve similarity over time. As shown in \tabref{tab:inter-session-pr}, TreeLoc achieved the highest mean recall, F1, and AUC as well as the lowest standard deviation ($\sigma$), demonstrating the temporal stability of the tree-based pipeline. Performance in \texttt{Karawatha} was slightly lower than in \texttt{Venman} due to there being fewer trees in open areas, yet higher F1 and AUC still indicated practical utility.

In comparison, BTC and MapClosure achieved lower scores because their local descriptors are affected by scene appearance, resulting in less reliable matches. SC++, SOLID, and RING++ often achieved better performance than local descriptor-based methods by compressing scenes, but exhibited larger standard deviations when temporal discrepancies were significant. As shown in \figref{fig:consistency}, the second-best performing method, SC++, retrieves many true positives on the same day between \texttt{V-01} and \texttt{V-02}, but failed to detect matches between \texttt{V-02} and \texttt{V-03} due to environmental change. Meanwhile, TreeLoc consistently retrieves correct matches across all time intervals, demonstrating the robustness of the tree-based descriptor to environmental changes.

\subsection{Localization Accuracy}
We compared pose estimation across four metrics on three sequences, including \texttt{K-04}, which had the lowest place recognition performance. Since some baselines only supported 3-DoF pose estimation, we reported both 3-DoF and 6-DoF results for TreeLoc to ensure fair comparison. SC++ was excluded from this evaluation as it only provides yaw angle prediction. An ablation study was also conducted on a variant of TreeLoc that used only the first step of our two-step geometric verification pipeline, in which triangle centroids $\mathbf{q}$ were used. This is referred to as TreeLoc-f.
As shown in \tabref{tab:localization}, TreeLoc outperformed the baselines on most metrics, achieving centimeter-level accuracy and high success rates, including \texttt{K-04}, even without ICP refinement. RING++'s BEV-based global descriptor was sensitive to viewpoint and resolution, resulting in large translation and rotation errors. MapClosure's ORB features lacked consistency across the forest, hindering robust matching. BTC used triangles and binary vectors, but performed worse than TreeLoc due to inconsistent observations of triangle vertices across different viewpoints and forest densities.

\begin{figure}[!t]
    \centering
    \includegraphics[width=.9\columnwidth, trim=4.5mm 4.5mm 4.5mm 0mm, clip]{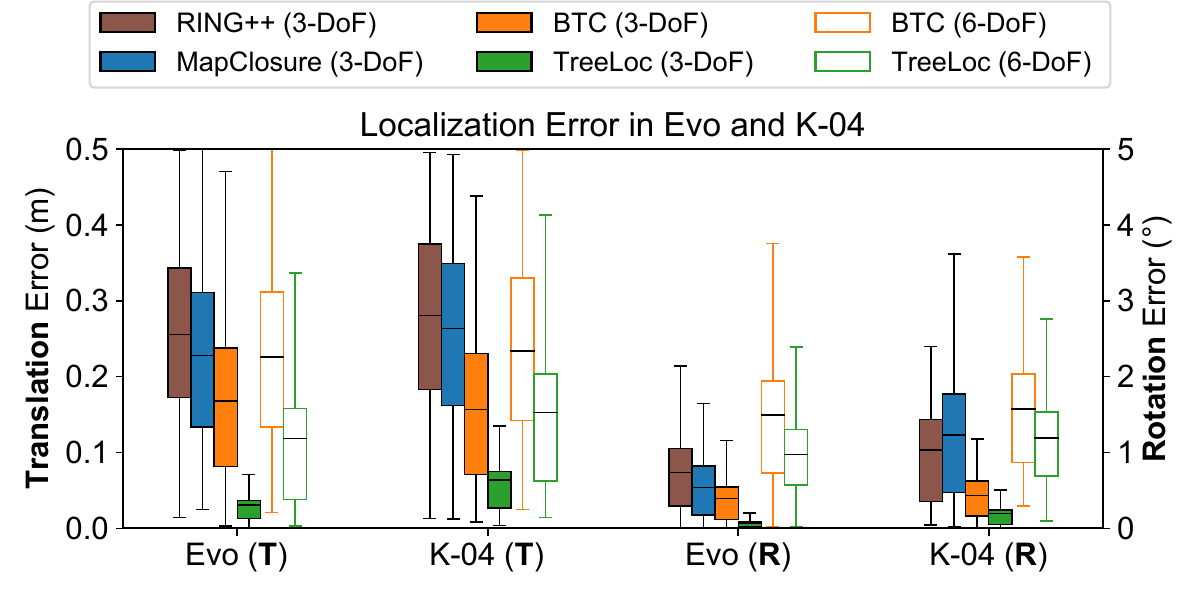}
    \vspace{-1.5mm}
    \caption{Localization errors for 3-DoF (Filled) and 6-DoF (Unfilled). TreeLoc shows lower errors and more stable results in both cases.}
    \label{fig:bargraph}
    \vspace{-7mm}
\end{figure}

In 6-DoF evaluations, BTC showed a significant drop in success rate, as tree height variations hindered relative $z$ estimation.
In contrast, TreeLoc maintained strong performance in both 3-DoF and 6-DoF, with lower median error and narrower interquartile range as shown in \figref{fig:bargraph}.
An ablation study showed that the first step in geometric verification already outperformed BTC’s 3D triangle matching, while the second step with stem centers further improved accuracy. These reflect TreeLoc’s 2D design, which removes height ambiguity, enables stable geometric matching. Moreover, combined with axis alignment and base height, it allows 6-DoF localization, benefiting practical applications in \secref{sec:applications}.

\begin{table}[]
\centering
\caption{Localization Performance (Up: 3-DoF, Down\(^{*}\): 6-DoF)}
\vspace{-1.5mm}
\label{tab:localization}
\resizebox{\columnwidth}{!}{%
\begin{tabular}{l|cccc|cccc}
\toprule
\multicolumn{1}{c|}{} &
  \multicolumn{4}{c|}{Oxford (\texttt{Stein})} &
  \multicolumn{4}{c}{Wild-Places (\texttt{K-04})} \\
Method &
  R@50 &
  SR &
  TE [m] &
  RE [\(\degree\)] &
  R@50 &
  SR &
  TE [m] &
  RE [\(\degree\)] \\ \midrule
RING++     & 0.227 & 0.388 & 0.266 & 1.031 & 0.257 & 0.409 & 0.281 & 1.031 \\ 
MapClosure & 0.092 & 0.806 & 0.198 & 0.795 & 0.411 & 0.544 & 0.255 & 1.089 \\ 
BTC        & 0.638 & 0.780 & 0.225 & 0.641 & 0.548 & \cellcolor[HTML]{ecf8f1}\textbf{0.961} & 0.157 & 0.431 \\
\cellcolor[HTML]{f3f7fc}\textbf{TreeLoc-f} & \cellcolor[HTML]{ecf8f1}0.953 & \cellcolor[HTML]{ecf8f1}0.982 & \cellcolor[HTML]{ecf8f1}0.100 & \cellcolor[HTML]{ecf8f1}0.248 & \cellcolor[HTML]{ecf8f1}0.557 & 0.765 & \cellcolor[HTML]{ecf8f1}0.112 & \cellcolor[HTML]{ecf8f1}0.370 \\ 
\cellcolor[HTML]{f3f7fc}\textbf{TreeLoc} & \cellcolor[HTML]{def3e6}\textbf{0.970} & \cellcolor[HTML]{def3e6}\textbf{0.987} & \cellcolor[HTML]{def3e6}\textbf{0.053} & \cellcolor[HTML]{def3e6}\textbf{0.137} & \cellcolor[HTML]{def3e6}\textbf{0.653} & \cellcolor[HTML]{ecf8f1}{0.895} & \cellcolor[HTML]{def3e6}\textbf{0.064} & \cellcolor[HTML]{def3e6}\textbf{0.197} \\ \midrule \midrule
BTC\(^{*}\)        & 0.290 & 0.398 & 0.236 & 1.953 & 0.414 & \cellcolor[HTML]{ecf8f1}{0.766} & 0.234 & 1.576 \\
\cellcolor[HTML]{f3f7fc}\textbf{TreeLoc-f\(^{*}\)} & \cellcolor[HTML]{ecf8f1}0.876 & \cellcolor[HTML]{ecf8f1}0.916 & \cellcolor[HTML]{ecf8f1}0.168 & \cellcolor[HTML]{ecf8f1}1.055 & \cellcolor[HTML]{ecf8f1}0.519 & 0.709 & \cellcolor[HTML]{ecf8f1}0.199 & \cellcolor[HTML]{ecf8f1}1.321 \\ 
\cellcolor[HTML]{f3f7fc}\textbf{TreeLoc\(^{*}\)} & \cellcolor[HTML]{def3e6}\textbf{0.908} & \cellcolor[HTML]{def3e6}\textbf{0.940} & \cellcolor[HTML]{def3e6}\textbf{0.126} & \cellcolor[HTML]{def3e6}\textbf{1.013} & \cellcolor[HTML]{def3e6}\textbf{0.609} & \cellcolor[HTML]{def3e6}\textbf{0.832} & \cellcolor[HTML]{def3e6}\textbf{0.153} & \cellcolor[HTML]{def3e6}\textbf{1.192} \\ \bottomrule
\end{tabular}%
}
\vspace{-3mm}
\end{table}

\subsection{Learning-based Methods in Forest Place Recognition}
We evaluated TreeLoc against learning-based methods on \texttt{Evo}, \texttt{Stein am Rhein}, and \texttt{1008-01} to highlight its algorithmic generality. For the baselines, we used checkpoints from ForestLPR, trained on the Wild-Places, and evaluated them in a zero-shot setting, with test sequences beyond the training domain. As shown in \tabref{tab:learning}, TreeLoc outperformed learning-based methods, which degraded under domain shifts in canopy height and data acquisition. Regardless of these shifts, TreeLoc ensures robust place recognition through stable stem extraction across diverse environments.

In \figref{fig:feature}(f), ForestLPR demonstrated strong true positive discrimination but often produced predictions that deviated from the ground truth, despite being designed for forests. The Precision–Recall curves in \figref{fig:pr_curve} further illustrate TreeLoc’s ability to maintain high precision at high recall, consistently outperforming learning-based baselines. Despite the small number of trees in \texttt{1008-01}, TreeLoc achieved a moderate R@1 and the highest F1 and AUC scores, highlighting its precision under low tree density. These results suggest that algorithmic approaches based on robust, domain-aligned features can offer superior generalization and efficiency compared to learning-based methods in natural environments.

\subsection{Ablation Studies}
We evaluated the TreeLoc configuration through ablations on \texttt{Stein am Rhein} and \texttt{V-04}, reporting R@1, F1, R@50, and runtime in \tabref{tab:ablation}.
The experiments isolated the contributions of TDH, DBH, and the alignment method.


\begin{table}[!t]
\centering
\caption{Performance Comparison with Learning-based Method}
\vspace{-1.5mm}
\label{tab:learning}
\resizebox{\columnwidth}{!}{%
\begin{tabular}{l|ccc|ccc|ccc}
\toprule
\multicolumn{1}{c|}{} &
  \multicolumn{3}{c|}{Oxford (\texttt{Evo})} &
  \multicolumn{3}{c|}{Oxford (\texttt{Stein})} &
  \multicolumn{3}{c}{Botanic (\texttt{1008-01})} \\
Method &
  R@1 &
  F1 &
  AUC &
  R@1 &
  F1 &
  AUC &
  R@1 &
  F1 &
  AUC  \\ \midrule
TransLoc3D & 0.613 & 0.760 & 0.814 & 0.579 & 0.725 & 0.760 & 0.760 & 0.863  & 0.877  \\ 
LoGG3D-Net & \cellcolor[HTML]{ecf8f1}0.703 & \cellcolor[HTML]{ecf8f1}0.826 & \cellcolor[HTML]{ecf8f1}0.872 & 0.593 & 0.744 & 0.642 & \cellcolor[HTML]{def3e6}\textbf{0.864} & \cellcolor[HTML]{ecf8f1}0.927 & \cellcolor[HTML]{ecf8f1}0.955 \\ 
MinkLoc3Dv2 & 0.424 & 0.595 & 0.475 & 0.409 & 0.581 & 0.497 & 0.558 & 0.717 & 0.749 \\ 
ForestLPR  & 0.672 & 0.804 & 0.839 & \cellcolor[HTML]{ecf8f1}0.628 & \cellcolor[HTML]{ecf8f1}0.771 & \cellcolor[HTML]{ecf8f1}0.799 & \cellcolor[HTML]{ecf8f1}0.825 & 0.904 & 0.923 \\
\cellcolor[HTML]{f3f7fc}\textbf{TreeLoc}       &  \cellcolor[HTML]{def3e6}\textbf{0.907} & \cellcolor[HTML]{def3e6}\textbf{0.966}& \cellcolor[HTML]{def3e6}\textbf{0.992} & \cellcolor[HTML]{def3e6}\textbf{0.871} & \cellcolor[HTML]{def3e6}\textbf{0.945} & 
\cellcolor[HTML]{def3e6}\textbf{0.988} & 0.773 & \cellcolor[HTML]{def3e6}\textbf{0.945} & \cellcolor[HTML]{def3e6}\textbf{0.980}\\ \bottomrule
\end{tabular}%
}
\vspace{-3mm}
\end{table}

\begin{table}[t]
\centering
\caption{Ablation Studies on TreeLoc (Alignment using Tree (\textbf{T}) and Ground (\textbf{G}); Time in milliseconds)}
\vspace{-1.5mm}
\label{tab:ablation}
\resizebox{\columnwidth}{!}{%
\begin{tabular}{c c c|cccc|cccc}
\toprule
\multicolumn{3}{c|}{{Configuration}} &
\multicolumn{4}{c|}{Oxford (\texttt{Stein})} &
\multicolumn{4}{c}{Wild-Places (\texttt{V-04})} \\ 
\# TDH & DBH & Align &
R@1 & F1 & R@50 & Time & 
R@1 & F1 & R@50 & Time \\ \midrule
\xmark   & \xmark & \textbf{T} & 0.842 &  0.930& \cellcolor[HTML]{ecf8f1}0.904 & 252.6 & 0.842 & 0.926 & \cellcolor[HTML]{ecf8f1}0.569 & 508.1 \\
50  & \xmark & \textbf{T} & 0.819 & 0.932 & 0.874 &  26.7 & 0.727 & 0.850 & 0.454 &  26.8 \\
50  & \cmark & \textbf{T} & \cellcolor[HTML]{ecf8f1}0.859 & 0.938 & 0.896 &  26.9 & 0.837 & 0.913 & 0.534 &  27.1 \\
100 & \xmark & \textbf{T} & 0.846 & 0.936 & 0.903 &  45.3 & 0.822 & 0.904 & 0.549 &  46.0 \\
100 & \cmark & \xmark & 0.748 & 0.878 & 0.701 &  45.5 & 0.825 & 0.904 & 0.469 &  47.4 \\
100 & \cmark & \textbf{G} & 0.856 & \cellcolor[HTML]{ecf8f1}0.940 &  0.888 &  45.7 & \cellcolor[HTML]{ecf8f1}0.882 & \cellcolor[HTML]{ecf8f1}0.937 & 0.566 &  45.9 \\
\cellcolor[HTML]{f3f7fc}100 & \cellcolor[HTML]{f3f7fc}\cmark & \cellcolor[HTML]{f3f7fc}\textbf{T} & \cellcolor[HTML]{def3e6}\textbf{0.871} & \cellcolor[HTML]{def3e6}\textbf{0.945} & \cellcolor[HTML]{def3e6}\textbf{0.908} &  45.3 & \cellcolor[HTML]{def3e6}\textbf{0.890} & \cellcolor[HTML]{def3e6}\textbf{0.942} & \cellcolor[HTML]{def3e6}\textbf{0.599} &  45.3 \\ 
\bottomrule
\end{tabular}%
}
\vspace{-6mm}
\end{table}

\textbf{Role of TDH:}
TDH is essential for reducing the candidate set during fine retrieval while maintaining, and often enhancing, place recognition performance. Without TDH, triangles formed at locations that overlap spatially but are distant from true positives can match in the hash space, resulting in inaccurate retrievals that are farther from the actual location. TDH reduces search time and stabilizes the matching process by establishing a location-centered distribution that prioritizes nearby candidates. As shown in \tabref{tab:ablation}, configurations using TDH achieved similar or higher R@1 and F1 score with substantially shorter runtime.

\textbf{Effect of DBH:}
To verify whether DBH provides meaningful information, we compared TDH with and without DBH. Applying DBH improved all metrics, with particularly notable improvements in R@1 and R@50. DBH enriches the representation by leveraging forest-specific cues and helps resolve ambiguity between similar geometric structures.

\textbf{Effect of Scene Alignment:}
Viewpoint variations cause roll and pitch distortions that affect the dual-descriptor used for place recognition. To evaluate the impact of our preprocessing step, we compared our proposed tree-based scene alignment with ground alignment (as used in ForestLPR) and a no-alignment variant that feeds raw centers directly to TDH and the triangle hash.
The results in \tabref{tab:ablation} demonstrate that using tree axis alignment yields a more stable inter-tree geometry, leading to consistent descriptor extraction and matching, and thereby improving all metrics. While ground alignment is beneficial on relatively flat terrain, it is less reliable on irregular terrain. Without alignment, roll and pitch incline distort in-plane distances and reduce triangle-key consistency, resulting in overall performance degradation.

\textbf{Runtime Analysis:}
We evaluated real-time feasibility by measuring the end-to-end time required for descriptor generation, candidate search, and pose estimation on an 11th Gen Intel Core i7-11700 @ 2.50 GHz. 
All TDH-enabled configurations achieve inference in under \unit{50}{ms}, enabling low-latency operation suitable for practical deployment.

\subsection{Application 1: Efficient Forest Map Management}
TreeLoc previously generated sets of reconstructed trees ($\mathcal{I}_t$) and dual-descriptors per scene using aggregated payloads (\(\mathcal{Z}_t\)). While this approach enabled localization at runtime, it lacked scalability due to duplicated storage and fragmented map management. To address this, we aggregate all observations into a single global tree database. At each location, descriptors are dynamically computed from nearby trees, removing the need for scene-specific processing.

To evaluate its efficiency, we compared storage requirements across three sequences collected at the same site as \texttt{Evo}: two missions in 2023 (\texttt{Mission01} and \texttt{Mission02}) using Hesai XT32, and one in 2025 (\texttt{Mission03}) using Hesai QT64. As shown in \tabref{tab:map_size}, the global tree database required only tens of kilobytes while encoding rich spatial information for each tree, including the stem axis, center, DBH, base height, and observation count. This representation was 97.8\% smaller than SC++ global descriptors and 99.9\% smaller than BTC local descriptors. Unlike prior experiments, TreeLoc does not store precomputed descriptors but generates them on demand in about \unit{1.4}{ms} per location, as shown in \tabref{tab:multi_session}. This is faster than BTC descriptor generation, which is slowed by processing heavy point clouds, and only marginally slower than loading such descriptors.

A key advantage of the global tree database is that its size scales with the number of unique trees rather than trajectory length, making it more suitable for long-term mapping. Moreover, its tree-level representation supports incremental updates: new missions can be aligned and integrated into the existing map. This is further demonstrated through multi-session alignment across temporally distinct sequences.

\begin{table}[!t]
\centering
\caption{Comparison of Storage Requirements for Localization}
\vspace{-1.5mm}
\label{tab:map_size}
\resizebox{1.0\columnwidth}{!}{%
\begin{tabular}{l|c|r|r|r|r}
\toprule
{Mission} & \# Scenes & Raw PCD & BTC Desc.& SC++ Desc. & \cellcolor[HTML]{f3f7fc}\textbf{TreeLoc} \\ \midrule
Mission01 & 452 & 1.7 GB & 478.1 MB & 3.6 MB & \cellcolor[HTML]{def3e6}\textbf{69.9 KB} \\
Mission02 & 788 & 3.0 GB & 857.6 MB & 6.5 MB & \cellcolor[HTML]{def3e6}\textbf{136.7 KB} \\
Mission03 & 222 & 0.2 GB & 241.2 MB & 1.8 MB & \cellcolor[HTML]{def3e6}\textbf{61.0 KB} \\ \bottomrule
\end{tabular}%
}
\vspace{-3mm}
\end{table}

\begin{table}[!t]
\centering
\caption{Multi-session Alignment Performance with Runtime}
\vspace{-1.5mm}
\label{tab:multi_session}
\resizebox{1.0\columnwidth}{!}{%
\begin{tabular}{l|c|c|c|c|c}
\toprule
Method & Time A [ms] & Time B [ms] & \# Matching Pairs & ATE [m] & ARE [$^\circ$] \\ 
\midrule
BTC (TP)   & \multirow{3}{*}{527.000} & \multirow{3}{*}{0.946} & 283  & 2.491 & 3.484 \\
BTC (SR)   &                          &                         & 73  & 6.085 & 13.378 \\
BTC (200) &                          &                         & 200  & 25.580 & 59.910 \\ \midrule

\cellcolor[HTML]{f3f7fc}\textbf{TreeLoc} & \cellcolor[HTML]{def3e6}\textbf{1.377} & \cellcolor[HTML]{def3e6}\textbf{-} & \cellcolor[HTML]{def3e6}\textbf{1110} & \cellcolor[HTML]{def3e6}\textbf{0.248} & \cellcolor[HTML]{def3e6}\textbf{0.492} \\
\bottomrule
\end{tabular}%
}
\small
\vspace{-0.1mm}
{$\,$}

\scriptsize 
\raggedright 
Time A: descriptor generation time, Time B: time to load descriptors from storage.
\vspace{-6mm}
\end{table}

\begin{figure}[!t]
    \centering
    \includegraphics[width=.9\columnwidth]{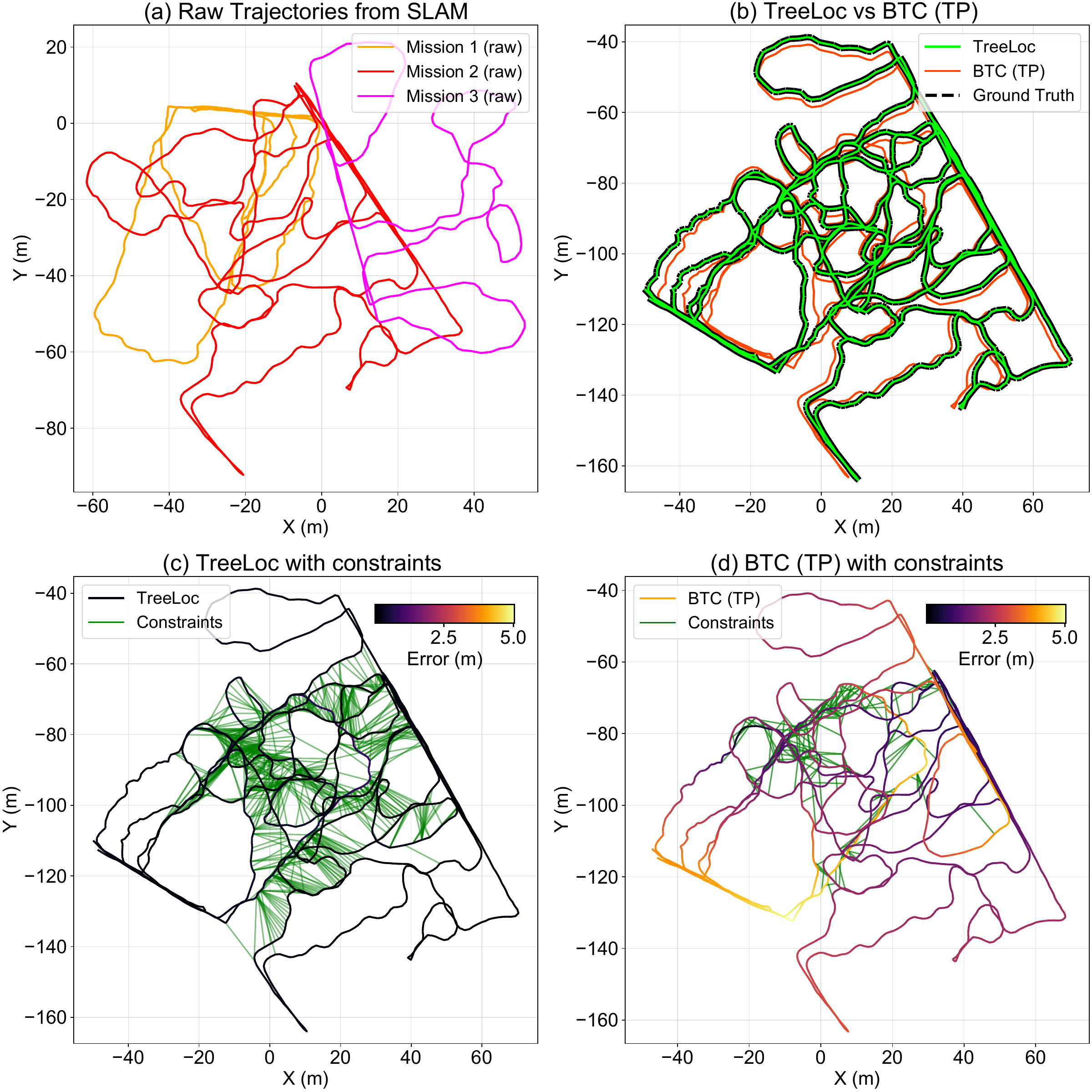}  
    \vspace{-2mm}
    \caption{(a) Initial trajectories before the pose graph optimization. (b) Alignment quality of TreeLoc and BTC (TP) with ground truth. TreeLoc shows tighter alignment. (c, d) Optimized trajectories, color-coded based on a translation error. Thanks to the diverse and reliable constraints, TreeLoc consistently reduces errors across the combined map, while BTC (TP) exhibits larger errors.}

    \label{fig:multisession}
    \vspace{-7mm}
\end{figure}

\subsection{Application 2: Lightweight Multi-session Alignment}
\label{sec:applications}
We conducted multi-session alignment across the three missions using the global tree database. Ground truth was obtained via ICP applied to manually matched scan pairs with spatial overlap. Since true positives are not known a priori, inter-session constraints were added only when the overlap ratio $\mathcal{O}(Q,C)$ exceeded 0.2. For BTC, we evaluated three variants: BTC~(TP), which retains only true positives; BTC~(SR), which retains successful localizations; and BTC~(200), which retains the top 200 matches based on feature similarity and reflects a realistic usage scenario. All constraints were inserted into a pose graph and optimized using GTSAM \cite{dellaert2012factor}, allowing loop closures to refine the SLAM trajectories \cite{wisth2022vilens}. The initial trajectories, which do not incorporate GPS measurements, are illustrated in \figref{fig:multisession}(a).

To ensure fairness, descriptors were computed from equivalent local information. TreeLoc queried nearby trees from the global tree database, while BTC extracted local crops from point clouds of full scans. As summarized in \tabref{tab:multi_session} and visualized in \figref{fig:multisession}(c), TreeLoc produced the lowest absolute translation error (ATE) and absolute rotation error (ARE), and the most valid constraints. The point cloud overlays in \figref{fig:Fig1}(d) and \figref{fig:Fig2} further demonstrate accurate alignment without relying on ICP refinement, confirming that the overlap ratio effectively filters false positives.
In contrast, BTC~(TP) preserved enough links for graph optimization but introduced larger translation errors, as seen in \figref{fig:multisession}(b) and (d). BTC~(SR) produced accurate matches but insufficient constraints for multi-session alignment. BTC~(200), while denser, suffered from false positives. Despite differences in LiDAR hardware and acquisition years, TreeLoc achieved consistent multi-session alignment. These results confirm the practicality of maintaining a single global tree database for compact, updatable, and accurate forest map management, as further illustrated in the accompanying video.

\section{Conclusion}
We present TreeLoc, a global localization framework for forests using tree stems as geometric primitives. It combines consistent 2D projection, TDH for coarse retrieval, and 2D triangle descriptors with geometric verification to estimate 6-DoF poses. TreeLoc outperforms \ac{SOTA} global and local descriptor-based methods across diverse datasets, demonstrating superior localization accuracy. Ablation studies confirm that TDH, DBH, and 2D projection based on tree axes enhance overall performance. Furthermore, a compact global tree database supports lightweight multi-session alignment and scalable map management. Although TreeLoc's reliance on tree stems may limit performance in open areas, it still outperforms all baselines across forested regions, including those with sparse trees, in most metrics. Future work will explore additional primitives, such as BEV contours, to enhance localization in open areas or environments with lower tree density. Overall, our results suggest that structure-aware, stem-centric geometry provides a robust and scalable foundation for long-term forest global localization.


\balance
\small
\bibliographystyle{IEEEtranN} 
\bibliography{string-short,references}

\end{document}